\documentclass[10pt,twocolumn,letterpaper]{article}

\usepackage{cvpr}
\usepackage{times}
\usepackage{epsfig}
\usepackage{graphicx}
\usepackage{amsmath}
\usepackage{amssymb}
\usepackage{multirow}
\usepackage[accsupp]{axessibility} % Improves PDF readability for those with disabilities.

\begin{document}

%%%%%%%%% TITLE
\title{Improving Face Recognition from Caption Supervision with Multi-Granular Contextual Feature Aggregation}

\author{Md Mahedi Hasan and Nasser Nasrabadi\\
West Virginia University\\
Morgantown, West Virginia, USA\\
{\tt\small mh00062@mix.wvu.edu, nasser.nasrabadi@mail.wvu.edu}
% For a paper whose authors are all at the same institution,
% omit the following lines up until the closing ``}''.
% Additional authors and addresses can be added with ``\and'',
% just like the second author.
% To save space, use either the email address or home page, not both
\and
%Nasser Nasrabadi\\
%Institution2\\
%First line of institution2 address\\
%{\tt\small secondauthor@i2.org}
}

\maketitle
\thispagestyle{empty}

%%%%%%%%% ABSTRACT
\begin{abstract}
We introduce caption-guided face recognition (CGFR) as a new framework to improve the performance of commercial-off-the-shelf (COTS) face recognition (FR) systems. In contrast to combining soft biometrics (\eg, facial marks, gender, and age) with face images, in this work, we use facial descriptions provided by face examiners as a piece of auxiliary information. However, due to the heterogeneity of the modalities, improving the performance by directly fusing the textual and facial features is very challenging, as both lie in different embedding spaces. In this paper, we propose a contextual feature aggregation module (CFAM) that addresses this issue by effectively exploiting the fine-grained word-region interaction and global image-caption association. Specifically, CFAM adopts a self-attention and a cross-attention scheme for improving the intra-modality and inter-modality relationship between the image and textual features, respectively. Additionally, we design a textual feature refinement module (TFRM) that refines the textual features of the pre-trained BERT encoder by updating the contextual embeddings. This module enhances the discriminative power of textual features with a cross-modal projection loss and realigns the word and caption embeddings with visual features by incorporating a visual-semantic alignment loss. We implemented the proposed CGFR framework on two face recognition models (ArcFace and AdaFace) and evaluated its performance on the Multi-Modal CelebA-HQ dataset. Our framework significantly improves the performance of ArcFace in both 1:1 verification and 1:N identification protocol.
\end{abstract}

%%%%%%%%%%%%%%%%%%%%%%%%%%%%%%%%%%%%%%%%%%%%%%%%%%%%%%%%%%%%%%%%%%%%%%
\section{Introduction}
Despite remarkable advancements in face recognition due to the adoption of margin-based loss functions~\cite{Deng2019, Kim_2022, magface}, face recognition in unconstrained scenarios remains a challenging problem~\cite{Yin2020}. The presence of covariate factors in an unconstrained environment, such as illumination, and pose variation, affects the face image quality, thus, decreasing the recognition performance. Providing auxiliary information, such as facial marks, ethnicity, and skin color, to a face recognition (FR) system can improve its recognition performance~\cite{Gonzalez_2018, Zhang_2015}. For example, video surveillance environment, where a prevalent commercially-off-the-shelf (COTS) system performs poorly~\cite{Yin2020, Zhu_2016}, the application of soft biometrics has been proven to improve the performance of hard biometrics~\cite{Tome_2014, Gonzalez_2018}.

Natural language captions that describe the visual contents of a face are an essential soft biometric trait for face recognition. Although captions are rich, they face many challenges that limit their application in the biometric system. As natural language contains high-dimensional information, it is often much more abstract than images. A short textual description of a given face consisting of a few sentences is insufficient to describe all the minute details of the facial features. Consequently, CGFR is significantly different from other tasks such as cross-modal image-text retrieval (ITR)~\cite{Lu2019, Li_2021_nips} and image-text matching (ITM)~\cite{Lee_2018}, where the matching text has a description of the various objects, background scenes, and styles, etc. 

To improve the performance of the FR system using CGFR, it is essential to find not only the semantic understanding of textual contents but also the proper association between visual and textual modalities. This is because the embedding space of images and text lies in different spaces due to the heterogeneity of the two modalities~\cite {Li_2021_nips}. Aligning the image features with word embeddings is thus crucial, as it has a significant impact on the performance of a cross-modal fusion algorithm~\cite {Li_2021_nips}. In this work, we finetune the state-of-the-art BERT model~\cite{Devlin2019} to update the contextual associations among words in the caption by incorporating a visual-semantic alignment loss~\cite{Xu_2018} and a cross-modal classification loss~\cite{Zhang_2018}. Finetuning the text encoder is essential because the BERT model was trained with different objectives than ours. Therefore, we finetune to achieve two objectives: (1) to learn visually aligned text embedding, \i.e., to realign word and caption embeddings with visual information, and (2) to enhance the discriminative power of textual features. 

However, a simple feature-level cross-modal fusion without fine-grained interaction between image-text tokens does not perform well. Therefore, we propose a novel module, namely, the contextual feature aggregation module (CFAM), to effectively carry out the fine-grained word-region interaction and global image-caption association on two different granularities. There are mainly three networks in the proposed CFAM: caption-level context modeling, word-level context modeling, and a feature aggregation network. Both context modeling networks adopt a self-attention and a cross-attention mechanism~\cite{Vaswani2017}. The self-attention mechanism increases the intra-modality relationship within each modality, while the cross-attention mechanism~\cite{Vaswani2017} improves the inter-modality relationship between image and textual features. The inputs to the feature aggregation network are the context-enhanced features from the word- and caption-level context modeling.

\begin{figure}[t]
	\centering
	\includegraphics[width=0.49\textwidth]{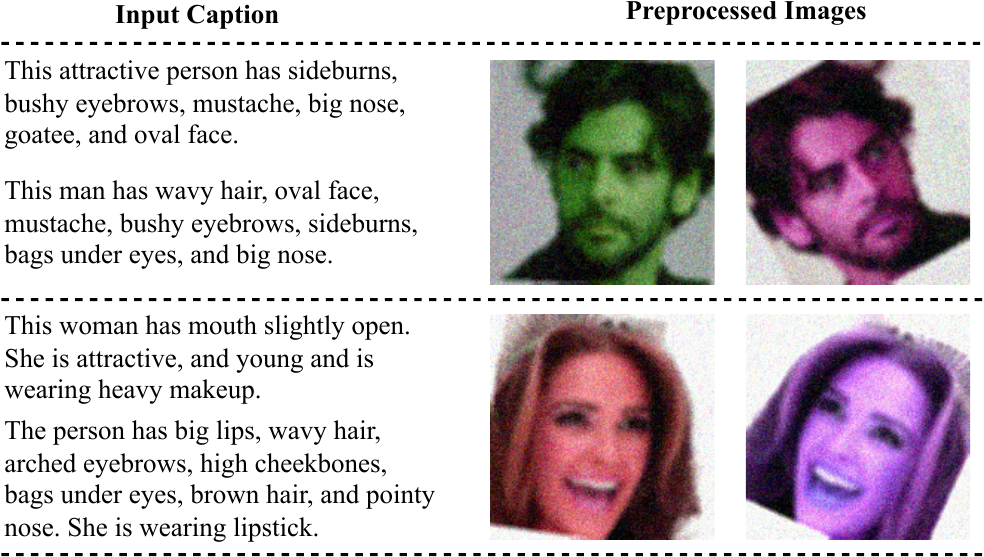}
	\caption{Sample image-caption pairs from the state-of-the-art Multi-Modal CelebA-HQ text-to-face dataset.}
	\label{fig:samples_dataset}
\end{figure}

We conduct our experiments on a benchmark text-to-face dataset, namely, Multi-Modal CelebA-HQ~\cite{Xia2021} (MMCelebA). Sample image-caption pairs of the dataset are illustrated in \figurename~\ref{fig:samples_dataset}. In fact, the dataset is based on a subset of the CelebA dataset~\cite{Liu_2015} that contains high-resolution images with very low variation. In our experiment, we remove the crucial face-alignment step and apply some pre-processing steps such as random sub-sampling, random rotation, horizontal flip to augment our database as well as downgrade the image quality in order to mimic real-world low-quality video surveillance scenarios. The verification rate of FR models, drops drastically on this preprocessed dataset because the images are corrupted with down-sampling and noise, which adversely affect their facial analysis procedure~\cite{Zhu_2016}. We then apply our CGFR framework to both models. The experimental results demonstrate a remarkable performance leap over these FR models.

%contributions
In this study, our contributions are: (1) exploring a new paradigm to improve face recognition with natural language supervision, (2) proposing the CFAM module to exploit fine-grained interaction among local and global features using word- and caption-level of granularities, (3) designing a textual feature refinement module (TFRM) to refine textual features and align them with visual content by fine-tuning the BERT encoder, and (4) conducting extensive experiments on the MMCelebA~\cite{Xia2021} dataset using the proposed CGFR framework to demonstrate substantial improvements over existing FR models. Finally, this work demonstrates excellent potential for caption-guided face recognition and provides a promising approach for further research.

% rest of the paper
The rest of this paper is organized as follows: an overview of the related works is presented in Section~\ref{sec:related_work}. A detailed description of the proposed method, including steps to finetune the BERT encoder, is presented in Section~\ref{sec:framework}. In Section~\ref{sec:expriments}, we demonstrate the experimental evaluation of the CGFR framework. Finally, we summarize our results with some possible future research directions in Section~\ref{sec:conclusion}.

%====================================================================================
\section{Related Work}\label{sec:related_work}
\subsection{Soft Biometrics}
Most of the prior works in the literature on improving face recognition using soft biometrics have been based on using categorical labels~\cite{Gonzalez_2018, Zhang_2015}. Zhang \etal~\cite{Zhang_2015} integrated a set of five soft biometrics (ethnicity, gender, eyebrow, eye color, and hair color) with hard biometric systems. Compared to the baseline recognition rates at FAR = 1e-3, their verification rate improved up to 15.5\% when introducing all the soft biometrics and 16.4\% when introducing gender information on the ugly part of the GBU database. Furthermore, authors in~\cite{Gonzalez_2018} empirically proved that a manual estimation of the six most discriminative soft biometrics improves the relative performance of the FR systems (COTS Face++ and VGG-face) up to 40\% over the LFW database.

\subsection{Caption-Guided Face Recognition}
Several early works have been proposed for caption-supervised face recognition~\cite{Huang_2020, Guillaumin_2012}. Huang \etal~\cite{Huang_2020} improved state-of-the-art face recognition using web-scale images with captions by learning the feature space in an iterative label expansion process. However, they only employed captions to extrapolate the labels of the face identity. 

Recently, with the development of generative adversarial networks (GANs)~\cite{Go14} and transformers~\cite{Vaswani2017}, text-to-face synthesis~\cite{Wang_2021, Sun_2022}, and facial attribute editing~\cite{He_2019, Xia2021} with textual descriptions have gained increasing popularity. For example, TediGAN~\cite{Xia2021} uses latent code optimization of pre-trained StyleGAN for caption-guided facial image generation and manipulation. In contrast to these works, we introduce a new line of research by exploring the idea of using natural language captions to improve the performance of the FR models. As there are no publicly available datasets that contain large-scale image-caption pairs for our task, we employ MMCelebA~\cite{Xia2021} dataset which has been widely used for text-to-face synthesis.

\subsection{Attention Techniques}
Recently, different attention mechanisms, such as self-attention, cross-attention, etc., have been extensively exploited in various multimodal tasks~\cite{Li_2017, Lee_2018, Xu_2018, Ye_2019}. Cross-attention or co-attention is an attention mechanism initially proposed in transformers~\cite{Vaswani2017} that interacts with two embedding sequences from different modalities (\eg, text, image). Li \etal~\cite{Li_2017} propose a latent co-attention mechanism in which spatial attention relates each word to corresponding image regions. Also, Lee \etal~\cite{Lee_2018} developed a stacked cross-attention network that learns the cross-modal alignments among all regions in an image and words in a sentence. Xu \etal~\cite{Xu_2018} applied an attention mechanism to guide the generator to focus on different words while generating various image sub-regions. They also proposed a deep attentional multimodal similarity model (DAMSM) to improve the similarity between the generated images and the given descriptions. To re-weight the importance of local image regions in tasks such as image synthesis~\cite{Xu_2018}, image caption generation~\cite{Xu_2015}, image segmentation~\cite{Shi_2018, Ye_2019}, and image-text matching~\cite{Lee_2018, Li_2017}, word-level attention has been employed. However, only employing word-level attention cannot ensure global semantic consistency due to the diversity of the text and image modalities. Global contextual information is also important as it provides more information on the visual content of the image and the context of the caption. It also drives the global features toward a semantically well-aligned context.

\subsection{Multimodal Representation Learning}
In recent years, dual-stream approaches, where the image and text encoder are trained on large-scale datasets individually with different cross-modality loss functions, have become widely popular in tackling various multimodal downstream tasks~\cite{Radford2021}. A lot of cross-modal loss functions such as contrastive~\cite{Radford2021, Li2021}, triplet~\cite{Lee_2018}, word-region alignment~\cite{Xu_2018}, cross-modal projection~\cite{Zhang_2018}, etc., have been proposed as part of the training objectives. Zhang \etal~\cite{Zhang_2018} proposed a novel projection loss that consists of two losses: a cross-modal projection matching (CMPM) loss for computing the similarity between image-text pairs and a cross-modal projection classification (CMPC) loss for learning a more discriminative visual-semantic embedding space. Also, Liao \etal~\cite{Liao_2021} proposed learning an optimal multimodal representation by maximizing the mutual information between the local features of the images and the sentence embedding. However, most of the dual-stream approaches in the literature cannot effectively and accurately exploit the fine-grained interaction among word-region features.

Furthermore, in other works, image and textual features, extracted from separate encoders, are often concatenated to feed into a fusion module to learn the joint representations~\cite{Ye_2019}. However, a simple fusion scheme may not be effective since the unaligned visual and word tokens lack prior relationships. Therefore, cross-modal interaction of local and global contexts is essential. For example, Niu \etal~\cite{Niu_2020} map phrases-region and image-caption into a joint embedding space using an image-text alignment method that consists of three different granularities: global-global alignment, global-local alignment, and local-local alignment.

In this work, we adopt a dual-stream approach to extract facial and textual features from pre-trained encoders. We apply a visual semantic alignment loss, known as DAMSM~\cite{Xu_2018}, to align the visual and word tokens locally and globally. We also employ CMPC loss~\cite{Zhang_2018} to enhance the discriminative power of the features. Finally, for fine-grained cross-modal interaction, we design CFAM. 

\begin{figure*}[t]
	\centering
	\includegraphics[width=0.84\textwidth]{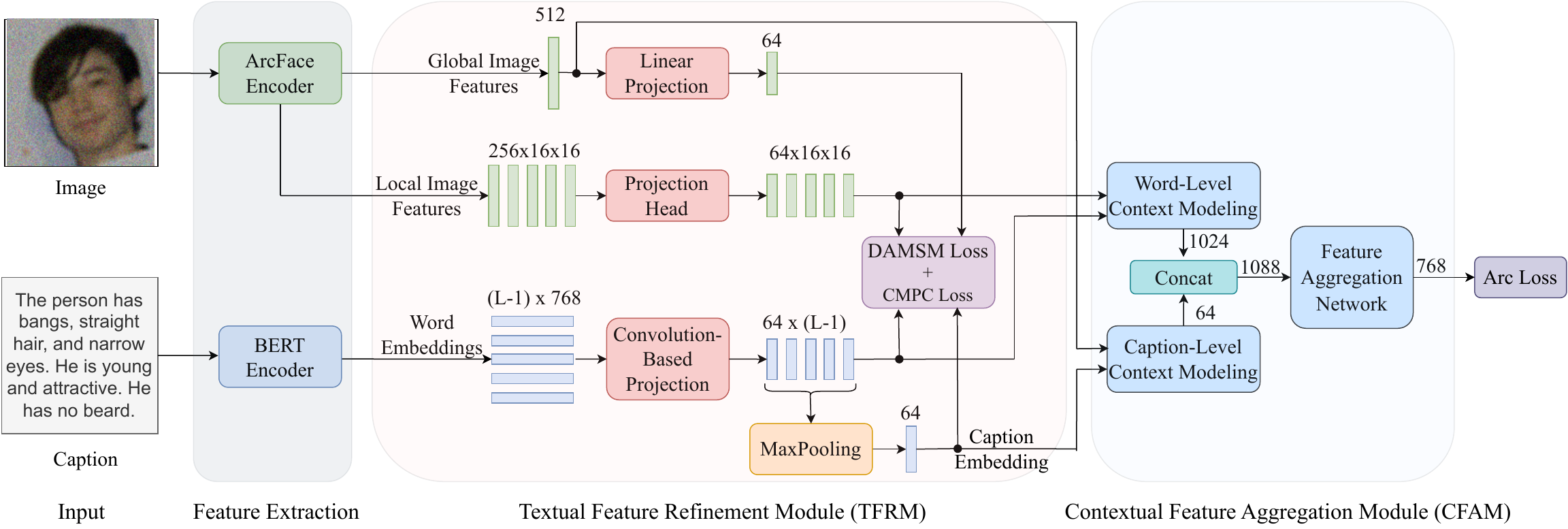}
	\caption{An overview of our proposed CGFR framework: it contains an ArcFace FR model and a pre-trained BERT encoder for extracting the facial features and textual embeddings from the input image-caption pair, respectively. First, TFRM updates contextual associations of the text embedding by finetuning the text encoder using the state-of-the-art DAMSM loss~\cite{Xu_2018} and a cross-modal projection loss~\cite{Zhang_2018}. Next, CFAM fuses the facial features with textual embeddings through cross-attention at both the word and caption-level of granularities.}
	\label{fig:network_arch}
\end{figure*}

%=====================================================================
\section{Framework}\label{sec:framework}
An overview of our proposed framework is depicted in \figurename~\ref{fig:network_arch}.

\subsection{Facial Feature Extraction}
We first employ the ArcFace model~\cite{Deng2019} as a feature extractor to extract the facial features from the input image. Specifically, we choose ResNet18-IR~\cite{He_2016, Deng2019} for the backbone of the ArcFace model, which was pre-trained on the  MS1MV3 dataset~\cite{Deng2019b}. Here, we modify the ResNet18-IR architecture by replacing the global average-pooling layer with a fully connected layer. The output of the fully connected layer is a 512-dimensional feature vector, which is considered as the global features $\textbf{v}~\epsilon~\mathbb{R}^{512}$ of the input image, as it contains high-level visual information. We extract the local features of the image $I~\epsilon~\mathbb{R}^{256\times14\times14}$ from the output of the third IR block. The size of the input image is $3\times112\times112$. We further employ CGFR on the AdaFace model~\cite{Kim_2022}. Here, the backbone, ResNet18-IR, is similar to the ArcFace model; however, it was pre-trained on the WebFace4M dataset~\cite{Zhu2021webface}. In contrast to ArcFace, the input is a BGR image.

\subsection{Textual Feature Extraction}
\subsubsection{BiLSTM}
Most of the works in the literature usually employ a long short-term memory network (LSTM)~\cite{hochreiter_1997} as an encoder to extract text embeddings from natural language descriptions~\cite{Xu_2015, Xu_2018}. Therefore, in this work, as a baseline, we apply a bidirectional LSTM (BiLSTM)~\cite{Schuster_1997} as a text encoder to extract semantic vectors from the input captions. The BiLSTM encoder encodes the input caption as a matrix of $W~\epsilon~\mathbb{R}^{L\times D}$. Here, $D$ denotes the dimension of the word vector, and $L$ denotes the maximum number of words in a caption. In our experiment, for the BiLSTM encoder, we consider a maximum of 18 words per caption, and the dimension of the word embedding is 256. Therefore, for an input caption of $L$ words, the word embeddings are [${\mathbf{w_1}, \mathbf{w_2}, \cdots \mathbf{w_L}}$], where $\mathbf{w_L}~\epsilon~\mathbb{R}^{D}$ is the caption embedding.

\subsubsection{BERT}
One of the limitations of traditional word embedding (such as word2vec) is that they only have one context-independent embedding for each word. Devlin \etal~\cite{Devlin2019} introduced BERT, a deep bidirectional encoder that considers the context of a word for each occurrence. In this work, we adopt a pre-trained BERT-base model~\cite{Devlin2019} with 12 encoder layers, each having 12 attention heads. It obtains the contextual embedding of each word by exploiting all the words in the caption. Furthermore, in addition to the input tokens, we add a classification token, $[CLS]$, at the beginning and a separator token, $[SEP]$, at the end of each sentence in the caption. The maximum length of the token sequence, $L$, is set to $21$. Additional padding tokens, $[PAD]$, are added for short-length captions after the last $[SEP]$ token. Extra tokens are truncated if the length of the input tokens is higher than $L$. Therefore, the input to the BERT-base model looks like this:
\begin{equation*}
{[CLS],~w_{2},~w_{3},\cdots,~w_{L-3},~[SEP],~[PAD],~[PAD],\cdots}
\end{equation*}

The output of the BERT layer gives a word matrix, $W~\epsilon~\mathbb{R}^{L\times768}$, where each contextualized token has an embedding of 768 dimensions. Here, the first token, $[CLS]$, is a classification token that represents the global embedding of the caption. The remaining $L-1$ tokens represent the contextualized word embeddings. In addition to BERT-base, we also experimented with other variants of BERT such as BERT-large, DistilBERT-base~\cite{Sanh_2019}, and RoBERTa-base~\cite{Liu2019}. However, in our experiments, we found that the performance of these variants is almost the same.

\subsection{Textual Feature Refinement Module}
In this subsection, we briefly describe the proposed textual feature refinement module. Because our text encoder was pre-trained with objectives that are totally different from ours and it creates embeddings that are unaligned to the image features, we need to refine the textual features. As shown in \figurename~\ref{fig:network_arch}, our TFRM consists of a convolution-based projection for text embeddings, a projection head for local image features, and a module to implement the visual-semantic alignment loss, DAMSM, and a cross-modal projection classification loss.

\subsubsection{Projection Heads}
\begin{figure}[t]
	\centering
	\includegraphics[width=0.50\textwidth]{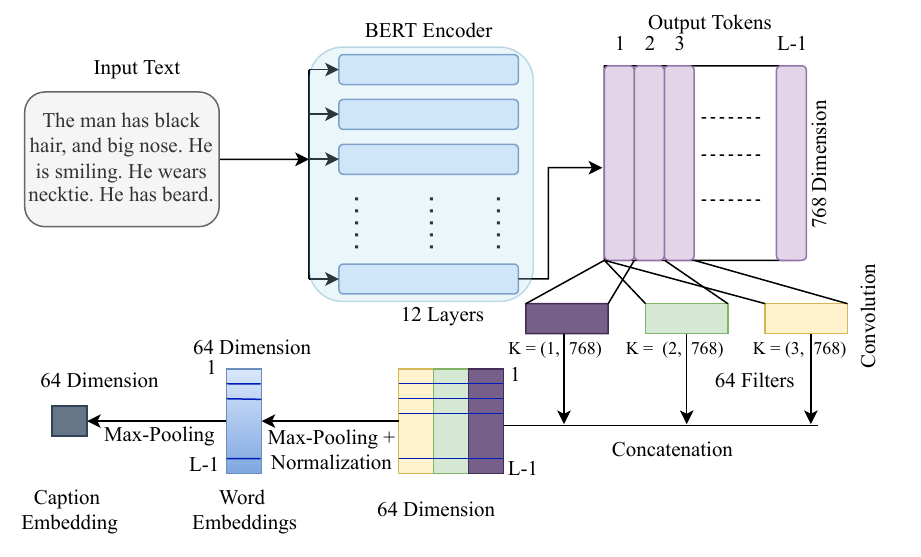}
	\caption{The proposed convolution-based projection for creating the word embeddings and global embedding for the input caption. 2D-convolutions with three different kernel sizes are applied to the output representations of the BERT encoder to extract both the word- and phrase-level information.}
	\label{fig:word_embeddings_projection}
\end{figure}

\paragraph{Convolution-based Projection:} As a caption has a natural order of word sequences, it is useful to extract not only word-level features but also phrase-level features. Thus, we apply a 2D-convolution to the output of the BERT sequence to extract both word-level and phrase-level information from the input caption. The first dimension of the kernel size $K$ is set to 1, 2, and 3 to project uni-gram, bi-gram, and tri-gram word sequences, respectively. For $K=$ 2 and 3, the word representations, $W$, are appropriately padded to maintain the fixed length of the sequence. All of these convolutions have a total of 64 filters with a stride of 1. Next, we apply the max-pooling operation followed by an $L_2$ normalization across the outputs of the convolutions to generate the word embeddings, $\mathbb{R}^{(L-1)\times64}$. \figurename~\ref{fig:word_embeddings_projection} illustrates the proposed scheme for creating word embeddings from the output of the BERT encoder.

\paragraph{Caption Embedding:} There are multiple ways of creating the global embedding for the input caption, $\textbf{c}~\epsilon~\mathbb{R}^{64}$. One common way to create caption embedding is to employ a linear projection followed by a batch normalization~\cite{Ioffe_15} on the $[CLS]$ token of the BERT output layer. We can also create the caption embedding by applying the max-pooling operation across the word embeddings of the convolution-based projection followed by an $L_2$ normalization. In our experiments, we achieved better results from the caption embedding which was created by the latter scheme. \figurename~\ref{fig:word_embeddings_projection} also depicts the scheme.

\paragraph{Projection Head for Image Features:} We project the local image features $I$ into a 64-dimensional space, which has also been empirically found to be the optimal dimension for the word embeddings. So, we design a projection head which consists of a $1\times1$ convolution with 64 filters and a Leaky ReLU activation~\cite{XU_2015_ReLU} for non-linearity.

\subsubsection{Objective Function}
\paragraph{DAMSM Loss:} AttnGAN~\cite{Xu_2018} introduced DAMSM loss to align image-caption pair by using word-level and caption-level attention. Let  $(W, I)$ denote an image-caption pair, where $W~\epsilon~\mathbb{R}^{L\times D}$ represents the word embeddings, and $I~\epsilon~\mathbb{R}^{H\times W\times D}$ represents the transposed local image features. Next, we apply DAMSM loss~\cite{Xu_2018} to perform cross-modal contrastive learning between image-caption pair. The loss actually minimizes the negative log posterior probability of the similarity scores between the image-caption pair.

\begin{figure*}[t]
	\centering
	\includegraphics[width=1.0\textwidth]{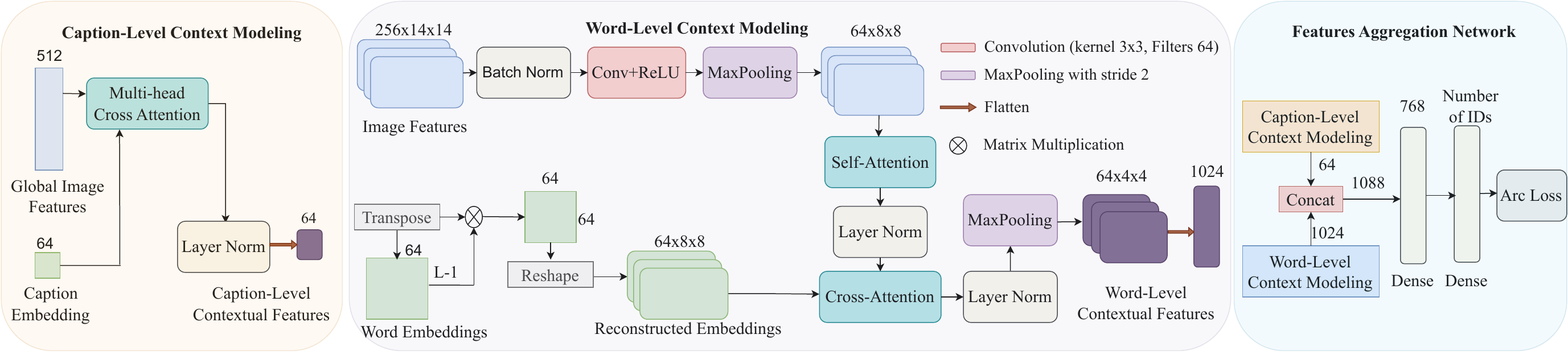}
	\caption{The block diagram of the proposed contextual feature aggregation module. It applies cross-modal feature interaction on both word and caption levels using a cross-attention mechanism. The module consists of three networks. The first network, caption-level context modeling, produces a 64-dimensional global context-enhanced features whereas the second network produces a 1024-dimensional regional context-enhanced features. The final network aggregates the contextual features and finds an optimal representation of it.}
	\label{fig:cfam_block_diagram}
\end{figure*}

\paragraph{Cross-Modal Projection Classification Loss:} In order to produce discriminative textual features, we also apply a cross-modal projection classification (CMPC) loss~\cite{Zhang_2018}. This loss first tries to project the representations from one modality onto the corresponding features from another modality and then classify them using normalized softmax loss. The input to the CMPC is the global image features, $\textbf{v}$, and caption embeddings, $\textbf{c}$. First, the image features are projected onto the normalized text embeddings, $\bar{\textbf{c}}$. Therefore, the normalized softmax loss for the image features, $L_{ipt}$, is given by:

\begin{equation}
\small 
\label{equ:loss_ipt}
L_{ipt} =\frac{1}{N}\sum_{i}-log(\frac{\exp(W^T_{y_i}\hat{\textbf{v}}_i)}{\sum_j{\exp(W^T_{j}\hat{\textbf{v}}_i)}}). 
\end{equation}
Here, $\hat{\textbf{v}}_i = \textbf{v}^T_i\bar{\textbf{c}_i}.\bar{\textbf{c}_i}$ denotes the vector projection of the image features. Now, let's project the text embeddings onto the normalized image features, $\bar{\textbf{v}}$. Therefore, the text classification loss function, $L_{tpi}$, is given by:

\begin{equation}
\small 
\label{equ:loss_tpi}
L_{tpi} =\frac{1}{N}\sum_{i}-log(\frac{\exp(W^T_{y_i}\hat{\textbf{c}}_i)}{\sum_j{\exp(W^T_{j}\hat{\textbf{c}}_i)}}).
\end{equation}
Here, $\hat{\textbf{c}}_{i} = \textbf{c}^{T}_{i}\bar{\textbf{v}}_i\bar{\textbf{v}}_i$ denotes the vector projection of the textual features. The total CMPC loss is the summation of the two losses, as defined by Eq.~\ref{equ:loss_ipt} and Eq.~\ref{equ:loss_tpi}.

\paragraph{Full Objective:} Our overall loss function is the weighted combination of the DAMSM and CMPC losses: 
\begin{equation}
\label{equ:obj_function}
L_{loss} = \lambda_{1}L_{DAMSM} + \lambda_{2}L_{CMPC},
\end{equation}
where, $\lambda_{1}$ and $\lambda_{2}$ are the hyperparameters that control the DAMSM and CMPC losses, respectively.

\subsection{Contextual Feature Aggregation Module}
In this study, we propose a contextual feature aggregation module (CFAM) that applies cross-modal feature interactions in two different granularities: word and caption. The block diagram of the proposed CFAM is illustrated in \figurename~\ref{fig:cfam_block_diagram}.

\subsubsection{Linear Fusion}
First, we concatenate the global image features $\textbf{v}~\epsilon~\mathbb{R}^{512}$ and the caption embedding $\textbf{c}~\epsilon~\mathbb{R}^{64}$ from the convolution-based projected head. Thus, we have a joint 576-dimensional multimodal representation. We then apply a fully connected (FC) layer. This network serves as a fusion scheme for our baseline approach.

\subsubsection{Word-Level Context Modeling}
In this network, we apply fine-grained cross-modal interactions between local image features and word embeddings. Here, we use word embeddings as cues to attend to the local image features extracted from the FR model. We also experimented with image features as cues to attend to words. However, that did not improve the performance, as words in a caption contain more abstract information than image regions. \figurename~\ref{fig:cfam_block_diagram} illustrates the word-level context modeling.

The inputs to the network are the word embeddings matrix, $W~\epsilon~\mathbb{R}^{L-1\times 64}$, and local image features $I~\epsilon~\mathbb{R}^{256\times14\times14}$. Batch normalization~\cite{Ioffe_15} is applied to the image features, before feeding it to a convolution layer of 64 filters with a kernel of size 3, and padding of 2. A max-pooling layer with a stride of size 2 is applied to the features map to reduce the spatial size to ${64\times8\times8}$. Next, a self-attention layer with a $scale = 0.5$ is applied to increase the intra-modality relationship among the local image features, followed by layer normalization~\cite{Ba_2016}. 

Thus, due to the application of self-attention, each image region now contains information about the whole image. In the self-attention layer, the $keys$, $queries$, and $values$ are learned from $1\times1$ convolutions. However, the number of filters in $1\times1$ convolutions for projecting  $key$ and $query$ are the multiplication of a $scale$ factor of the number filters of the $1\times1$ convolution for learning $values$. Note that the application of normalization and self-attention in this network, as analyzed in Table~\ref{table:abs_cfam}, is very crucial.

Contrary to image features, word embeddings $W$, have different dimensions. Therefore, we, first, calculate the correlation of the word embeddings matrix, $W^TW~\epsilon~\mathbb{R}^{64\times 64}$. Next, we reshape the embeddings matrix to size ${64\times8\times8}$. Similar to image features, we also experimented to implement self-attention to the reconstructed word features, but that does not improve the performance. The reason for this could be that as textual features are extracted from transformer-based BERT architectures, the intra-modal relationship among the features is already high. Afterward, the word embeddings and image features are fed into a cross-attention scheme to increase the inter-modality relationship. Here,  the $queries$ are learned from the word embeddings matrix, and $keys$ and $values$ are learned from the image features using $1\times1$ convolutions with a $scale$ of 0.5. Finally, after applying another max-pooling layer, we flatten the feature matrix to produce a 1024-dimensional output.

\subsubsection{Caption-Level Context Modeling}
Similar to the word level of granularity, we take the caption embedding as cues to attend to the global image features. Multi-head cross-attention~\cite{Vaswani2017} has been employed to explore inter-modal associations between global image features and caption embedding. First, we reshape the global image features into size $8\times 64$, $\textbf{v}~\epsilon~\mathbb{R}^{8\times64}$. Then, we calculate the $queries$ from caption embedding, $\textbf{c}~\epsilon~\mathbb{R}^{64}$ and the $keys$, and $values$ from global features $\textbf{v}$ using linear transformation. The total number of heads is 4. The output of the multi-head cross-attention is a 64-dimensional vector, which is followed by a layer normalization~\cite{Ba_2016}.

\subsubsection{Features Aggregation Network}
At the final stage of CFAM, we aggregate the contextualized features from the word-level CM and caption-level CM. A dense layer learns the optimal representation in a joint multimodal feature space. In our experiment, we found that the optimal dimension of the dense layer is 768. We also experimented to implement the CFAM module on the textual features extracted from the BiLSTM text encoder. However, it does not perform well as BiLSTM encoder does not produce contextual embeddings for word tokens.

\subsection{Training Strategy}
We train our proposed framework in two phases. First, we train the TFRM module to update the contextual embeddings of the BERT encoder using the objective function mentioned in Equation~\ref{equ:obj_function}. We finetune the BERT encoder for only 4 epochs and use a mini-batch AdamW optimizer~\cite{Loshchilov_2019} with a weight decay of 0.02. The learning rate is initialized to $0.00001$ and is warmed up to $0.0001$ after 2,000 training iterations. We then decrease it using the cosine decay strategy~\cite{Loshchilov_ICLR2017} to $0.00001$. The batch size is set to 16. For the projection head of both visual and textual streams, we employ the Adam optimizer~\cite{Kingma_2014} with $\beta_{1}=0.5$ and $\beta_{2}=0.99$. The initial learning rate, in this case, is set to $0.001$. In the second phase, we train the whole framework end-to-end for 24 more epochs. However, the text encoder and the projection head were trained with a similar setup to the first phase except with a lower learning rate. Note that, in all the phases, the parameters of the FR model were fixed.

%%%%%%%%%%%%%%%%%%%%%%%% ArcFace Evaluation %%%%%%%%%%%%%%%%%%%%%%%%%%%
\begin{table}[t]
	\centering
	\small
	\setlength{\tabcolsep}{2pt}
	\caption{The 1:1 verification and 1:N identification (Rank-1) results of our CGFR framework with ArcFace  trained on the MMCelebA dataset. The top row represents the results of ArcFace when pre-trained on MS1MV3 dataset~\cite{Deng2019b}. \label{table:baseline_arcface}}
	
	{\begin{tabular*}{19.5pc}{@{\extracolsep{4pt}}c| cc cc c@{}}\hline  \rule{0pt}{2ex}		
			\multirow{2}{*}{Architectures} &\multicolumn{2}{c}{ROC Curve} &\multicolumn{2}{c}{TPR@FPR} &Id(\%)  \\\cline{2-3}\cline{4-5} \cline{6-6}\rule{0pt}{2ex}
			
			&AUC &EER &1e-4 &1e-3 &Rank-1 \\ \hline\rule{0pt}{2ex}
			
			% original
			Pre-trained ArcFace~\cite{Deng2019}  &85.27 &23.48 &16.75 &25.73 &17.56\\\rule{0pt}{2ex}
			
			% first baseline
			Baseline &93.98 &13.50 &21.92 &31.28 &38.78\\\rule{0pt}{2ex}
			
			% proposed method
			CGFR &\textbf{98.51} &\textbf{6.65} &\textbf{62.83} &\textbf{63.28} &\textbf{57.65} \\\hline
	\end{tabular*}}{} 
\end{table}

\begin{figure}[t]
	\centering
	\includegraphics[width=0.48\textwidth]{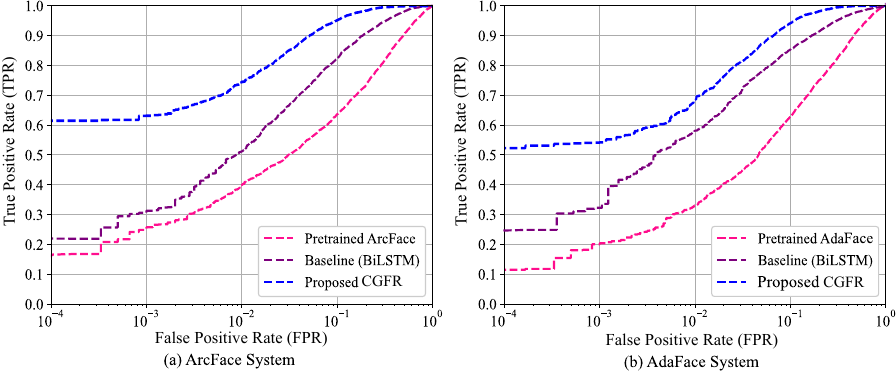}
	\caption{ROC curves of 1:1 verification protocol of the proposed CGFR framework with (a) ArcFace, (b) AdaFace FR models.}
	\label{fig:arc_ada_face_evaluation}
\end{figure}
%%%%%%%%%%%%%%%%%%%%%%%% End %%%%%%%%%%%%%%%%%%%%%%%%%%%

%===========================================================
\section{Experiments}  \label{sec:expriments}
\subsection{Dataset} \label{sub_sec:datasets}
The Mulit-Modal CelebA-HQ~\cite{Xia2021} (MMCelebA) is a large-scale text-to-face dataset, originally built for face image generation and facial attributes editing. It has a total of 30,000 high-resolution face images from the CelebA-HQ dataset~\cite{Liu_2015}. Each image has 10 auto-generated captions from a total of 38 facial attributes.

\subsection{Preprocessing}
First, we implement standard data augmentation techniques such as random sub-sampling, color jittering, horizontal flipping, rotation, and Gaussian noise to degrade the image quality of the MMCelebA dataset. Then, we resize all the images to 3 x 112 x 112. Sample preprocessed images are shown in the \figurename~\ref{fig:samples_dataset}.  The top row of Table~\ref{table:baseline_arcface} and Table~\ref{table:baseline_adaface} represent the performance of pre-trained ArcFace and AdaFace models on this preprocessed set, respectively.

\subsection{Implementation}
We implemented our architecture using two NVIDIA Titan RTX GPUs. In our experiment, we empirically set the hyper-parameters in Equations~\ref{equ:obj_function} as follows: $\lambda_{1}$ = 1, and $\lambda_{2}$ = 0.5. Since we employ pre-trained encoders, training the proposed framework is very fast. Finetuning pre-trained BERT model for 4 epochs takes approximately 80 minutes on the MMCelebA dataset while training the whole network end-to-end takes 8 hours. Also, due to the parallel strategy of our proposed framework, the model has a very low time complexity during inference. The inference time, which requires only one forward process, is 220ms for an image-caption pair.

%%%%%%%%%%%%%%%%%%%%%%%% Adaface Evaluation %%%%%%%%%%%%%%%%%%%%%%%%%%%
\begin{table}[t]
	\centering
	\small 
	\setlength{\tabcolsep}{1.5pt}
	\caption{The 1:1 verification and 1:N identification (Rank-1) results of our CGFR framework with AdaFace trained on the MMCelebA dataset. The top row represent the results of AdaFace when pre-trained on the WebFace4M dataset~\cite{Zhu2021webface}. \label{table:baseline_adaface}}
	
	{\begin{tabular*}{19.5pc}{@{\extracolsep{4pt}}c| cc cc c@{}}\hline  \rule{0pt}{2ex}			
			\multirow{2}{*}{Architectures} &\multicolumn{2}{c}{ROC Curve} &\multicolumn{2}{c}{TPR@FPR} &Id(\%)  \\\cline{2-3}\cline{4-5} \cline{6-6} \rule{0pt}{2ex}
			
			&AUC &EER &1e-4 &1e-3 &Rank-1 \\ \hline\rule{0pt}{2ex}
			
			% original
			Pre-trained AdaFace~\cite{Kim_2022} &85.55 &22.88 &11.46 &20.00 &8.45\\\rule{0pt}{2ex}
			
			% first baseline
			Baseline &93.97 &12.88 &24.28 &33.00 &22.55 \\\rule{0pt}{2ex}
			
			% proposed method
			CGFR &\textbf{98.10} &\textbf{7.52} &\textbf{53.08} &\textbf{54.12} &\textbf{43.23} \\\hline
	\end{tabular*}}{} 
\end{table}
%%%%%%%%%%%%%%%%%%%%%%%% End %%%%%%%%%%%%%%%%%%%%%%%%%%%

%===========================================================
\subsection{Performance Evaluation}
\paragraph{ArcFace System:} We compare our proposed CGFR to the pre-trained ArcFace and the baseline approach, as shown in Table~\ref{table:baseline_arcface}. Our baseline is a dual-stream model employing a BiLSTM text encoder with a linear fusion. We evaluated the CGFR model on one-to-one matching (1:1), wherein a face probe is compared to a single gallery image for verification, and one-to-many matching (1:N), wherein a face probe is compared to all gallery images for identification. In the 1:1 verification protocol, the proposed CGFR achieves the highest verification rates (VR). It improved the performance of pre-trained ArcFace by 71.68\% and the baseline by 50.74\% on the equal error rate (EER) metric. Also, on the true positive rate (TPR) and false positive rate (FPR) metrics, as illustrated in \figurename~\ref{fig:arc_ada_face_evaluation}(a), our proposed CGFR improves the VR(\%) by a significant margin. In particular, as compared to the pre-trained ArcFace model, our framework boosts TPR(@FPR=1e-4) from 16.75\% to 62.83\%.

Similarly, when compared to the baseline approach, our framework improves the TPR(@FPR=1e-4) from 21.92\% to 62.83\% and TPR(@FPR=1e-3) from 31.28\% to 63.28\%. Furthermore, in the 1:N identification protocol, the proposed CFGR secures an improvement of 48.66\% and 228.30\% on Rank-1 identification accuracy over baseline and pre-trained ArcFace, respectively. Therefore, as the results show, the ArcFace FR model, which performs poorly due to low quality and noise, could be significantly improved using natural language supervision.

\paragraph{AdaFace System:} In Table~\ref{table:baseline_adaface}, we conduct further experiments to evaluate the performance of our CGFR framework with an AdaFace FR model. As illustrated in \figurename~\ref{fig:arc_ada_face_evaluation}(b), our framework significantly improves the VR(\%) over the baseline and pre-trained AdaFace model. It improves the performance of pre-trained AdaFace by 67.13\% and the baseline by 41.61\% on the EER metric. Also, in 1:1 verification protocol, under the evaluation metric of TPR(@FPR=1e-4), our framework boosts the performance of pre-trained AdaFace from 11.46\% to 53.08\% and TPR(@FPR=1e-3) from 20.00\% to 54.12\%. Furthermore, as reported in Table~\ref{table:baseline_adaface}, the Rank-1 identification accuracy of our CGFR framework improves by 91.71\% over the baseline. Thus, the VR (\%) of the above-mentioned experiments proves the effectiveness and generalizability of the proposed framework.

\subsection{Analysis of CFAM}
\begin{table}[t]
	\centering
	\setlength{\tabcolsep}{4pt}
	\small 
	\caption{Ablation experiments of different networks on the CFAM module. Experimental results verifies the notion of fusing cross-modal features at multiple granularities improves 1:1 VR(\%).\label{table:abs_cfam}}
	
	{\begin{tabular*}{18pc}{c cc cc}\hline  \rule{0pt}{2ex}
			\multirow{2}{*}{Modules}   &\multicolumn{2}{c}{ROC Curve}   &\multicolumn{2}{c}{TPR@FPR} \\\cline{2-3}\cline{4-5}\rule{0pt}{2ex}
			                   &AUC &EER &1e-4 &1e-3 \\ \hline\rule{0pt}{2ex}
			w/o modules        &89.96  &18.27 &15.95 &21.42 \\ \hline\rule{0pt}{2ex}
			Word (w/o Norm)    &86.36 &22.27 &8.63 &20.63 \\\rule{0pt}{2ex}
			Word (w/o SA)      &96.30 &10.42 &27.02 &33.13\\\rule{0pt}{2ex}
			Word (SA+Norm)     &96.86 &9.88 &53.42 &54.45 \\\hline\rule{0pt}{2ex}
			Word + Caption     &97.22&9.38 &60.0 &60.75 \\
			Word + Caption + FAN &\textbf{98.51} &\textbf{6.65} &\textbf{62.83} &\textbf{63.28}\\\hline
	\end{tabular*}}{} 
\end{table}

\begin{figure}[t]
	\centering
	\includegraphics[width=0.36\textwidth]{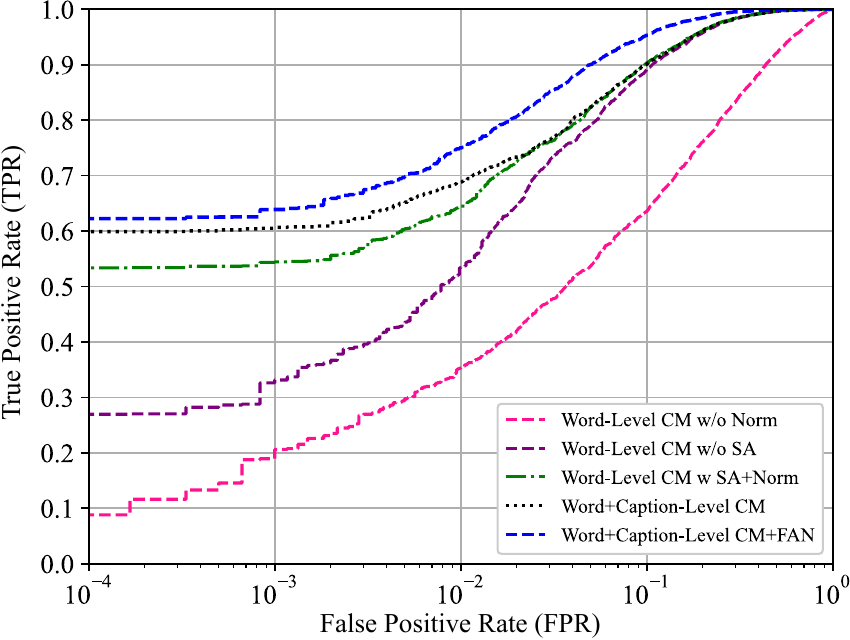}
	\caption{Face verification evaluation on different modules of the proposed CFAM using ROC curves.}
	\label{fig:abs_cfam}
\end{figure}

We design an ablation experiment to evaluate the effectiveness of the proposed CFAM module. Specifically, we analyze the role of individual granularities and attention schemes. Table~\ref{table:abs_cfam} demonstrates that a fusion scheme without any granularity decreases the VR(\%). Thus it proves the need for fusing contextual features at multiple granularities. In fact, under the evaluation metric of TPR, word-level contextual modeling (CM) increases the performance from 15.95\% to 53.42\% at FPR=1e-4 over the simple concatenation of multimodal features. However, the choice of adding normalization~\cite{Ioffe_15, Ba_2016} and self-attention are crucial to the performance of this module. We observe a drastic performance drop of 12.16\% in AUC without normalization (one batch norm~\cite{Ioffe_15} and two layer norm layer~\cite{Ba_2016}). Also, adding self-attention to the image features reduces the EER from 10.42\% to 9.88\%. 

We also observe that the fusion of word-level and caption-level CM improves the VR(\%) by 5.33\% on EER and 12.32\% on TPR@FPR=1e-4 compared to word-level CM. Furthermore, the ablation study shows that the implementation of the feature aggregation network further boosts the VR(\%), improving TPR from 60.0\% to 62.83\% (@FPR=1e-4). \figurename~\ref{fig:abs_cfam} depicts the performance comparison of these networks on ROC curves. Figure~\ref{fig:abs_cfam} illustrates that the proposed CFAM, with both CM networks and the feature aggregation network, achieves the highest VR(\%), proving the effectiveness of applying fine-grain word-region interaction and image-caption associations.

\subsection{Future Work}
In this work, we tested CGFR on distorted, low-quality images to mimic surveillance scenarios and evaluated the impact of the caption on two state-of-the-art FR models. Overall, we achieved a remarkable performance leap in this setup. In the future, we plan to evaluate our model on various resolutions and qualities of input images. Additionally, we aim to employ large-scale image-caption pair datasets to assess the generalizability of our proposed method. 

%===========================================================
\section{Conclusion}\label{sec:conclusion}
We have introduced a new framework, called the caption-guided face recognition (CGFR) model, to improve the performance of FR models using captions. Our framework is based on a dual-stream model with a TFRM, and a CFAM modules. CFAM applies fine-grained cross-modal feature interaction at multiple granularities using cross-attention. In contrast, TFRM helps the framework to learn an effective joint multimodal embedding space by  realigning the text embeddings with visual features. Our CGFR has significantly improved the performance of state-of-the-art FR models. It has also enhanced the robustness and reliability of the FR models by offering higher resistance to spoofing attacks. Since, only manipulating the face becomes insufficient to deceive the CGFR model, as the captions also provide discriminative and unique information about the face image.

\section{Acknowledgment}
This material is based upon a work supported by the Center for Identification Technology Research and the National Science Foundation under Grant 1650474.

{\small
\bibliographystyle{ieee}
\bibliography{IEEEabrv, ijcb.bib}
}

\end{document}